\documentclass[conference]{IEEEtran}
\usepackage{cite}
\usepackage{amssymb}
\usepackage{algorithmic}
\usepackage{graphicx}
\usepackage{balance}
\usepackage{url}
\usepackage{color}
\usepackage{xcolor}
\usepackage{multirow}
\usepackage{caption}
\usepackage{enumitem}
\usepackage{multicol}
\usepackage{subfloat}
\usepackage{pifont}
\graphicspath{ {Images/} }
\usepackage{subfig,epsfig,amsfonts,amsmath,cases,amssymb,graphics, latexsym, times, algorithmic, algorithm, bbm, color, soul,wrapfig,commen t}
\usepackage{epstopdf}
\usepackage{stfloats}
\usepackage{epsfig}
\usepackage{soul}
\usepackage{geometry}

\usepackage{textcomp}

\definecolor{aqua}{rgb}{0.7,0,0.7}

\def\BibTeX{{\rm B\kern-.05em{\sc i\kern-.025em b}\kern-.08em
    T\kern-.1667em\lower.7ex\hbox{E}\kern-.125emX}}

\begin{document}

\title{Adversarial Attacks to Machine Learning-Based Smart Healthcare Systems}
\author {\IEEEauthorblockN{ AKM Iqtidar Newaz\textsuperscript{\textdagger}, Nur Imtiazul Haque\textsuperscript{$\dagger\dagger$}, Amit Kumar Sikder\textsuperscript{\textdagger}, \\Mohammad Ashiqur Rahman\textsuperscript{$\dagger\dagger$}, and A. Selcuk Uluagac\textsuperscript{\textdagger}}
\textit{\textsuperscript{\textdagger}Cyber-Physical Systems Security Lab, \textsuperscript{$\dagger\dagger$}Analytics for Cyber Defense Lab} \\
Department of Electrical and Computer Engineering\\
Florida International University, Miami, USA\\
\{anewa001, nhaqu004, asikd003, marahman, suluagac\}@fiu.edu}

\vspace{-0.2in}

\maketitle

\begin{abstract}
The increasing availability of healthcare data requires accurate analysis of disease diagnosis, progression, and realtime monitoring to provide improved treatments to the patients. In this context, Machine Learning (ML) models are used to extract valuable features and insights from high-dimensional and heterogeneous healthcare data to detect different diseases and patient activities in a Smart Healthcare System (SHS). However, recent researches show that ML models used in different application domains are vulnerable to adversarial attacks. In this paper, we introduce a new type of adversarial attacks to exploit the ML classifiers used in a SHS. We consider an adversary who has partial knowledge of data distribution, SHS model, and ML algorithm to perform both targeted and untargeted attacks. Employing these adversarial capabilities, we manipulate medical device readings to alter patient status (disease-affected, normal condition, activities, etc.) in the outcome of the SHS. Our attack utilizes five different adversarial ML algorithms (HopSkipJump, Fast Gradient Method, Crafting Decision Tree, Carlini \& Wagner, Zeroth Order Optimization) to perform different malicious activities (e.g., data poisoning, misclassify outputs, etc.) on a SHS. Moreover, based on the training and testing phase capabilities of an adversary, we perform white box and black box attacks on a SHS. We evaluate the performance of our work in different SHS settings and medical devices. Our extensive evaluation shows that our proposed adversarial attack can significantly degrade the performance of a ML-based SHS in detecting diseases and normal activities of the patients correctly, which eventually leads to erroneous treatment.

\end{abstract}


\section{Introduction}\label{sec:intro}
Each year the healthcare industry is generating data at a staggering rate that is 
expected to reach 1,656 zettabytes by 2025 \cite{aml1}. This healthcare data have clinical, financial, and operational value if analyzed properly to extract important features. In this context, Machine Learning (ML) can be an effective tool for managing healthcare data. Recently, several ML-based healthcare applications have been proposed to perform different healthcare functionalities, such as disease detection, early diagnosis, treatment plan, and antidote discovery~\cite{esteva2019guide, newaz2020heka}. The integration of ML to identify and analyze clinical parameters has surely improved both the efficiency and quality of healthcare. The nature of feature extraction from the dataset of interest in ML can lead to patient-specific treatments and support, which can eventually lead to a reduction in medical costs and establish a better patient-doctor relationship. The global market of ML solutions in the healthcare sector is projected to reach \$34 billion by 2025 \cite{aml2}.


ML plays a vital role in the healthcare industry, such as the development of the new medical procedure, processing patient data, and the treatment of chronic diseases. In parallel to this progress, ML models exhibit unpredictable and overly confident behavior outside of the training distribution, and adversarial examples are a subset of this broader problem. The discovery of adversarial examples has exposed vulnerabilities in the state-of-the-art ML systems \cite{goodfellow2014explaining, abusnaina2019adversarial, shahriar2020g}. An adversarial example is an input engineered to cause misclassification in the ML algorithms. Indeed, adversarial ML has gained much popularity in the healthcare domain because of the possible limitations of the current ML models. For instance, an adversary may add new adversarial data to a healthcare ML model to falsely classify a hypothyroid patient \cite{mozaffari2014systematic}. Researchers also reported several adversarial attacks against ML model in medical image processing to alter the results by adding noises and misclassify a benign mole as malignant with high confidence~\cite{finlayson2019adversarial, newaz2020survey}. However, the healthcare system is aggressively adapting ML model to improve disease detection and patient treatment without explicitly addressing these reported adversarial attacks. 


In this paper, we present a new type of adversarial attacks that detects the pitfalls of a ML-based application in a smart healthcare system (SHS). In a SHS, multiple medical devices can be connected with each other to share the patient's vitals and detect normal and disease-affected activities by correlating various body functions. Our attack targets the underlying ML model used in a SHS for threat detection, disease recognition, and normal activity identification. In our attack, an adversary utilizes state-of-the-art adversarial attacks (HopSkipJump, Fast Gradient Method, Carlini \& Wagner, Decision Tree, Zeroth Order Optimization) against the ML-based SHS to perform both white-box and black-box attacks~\cite{chakraborty2018adversarial}. We show that the proposed attack can successfully misclassify the patient's state and manipulate the outcome to a specific state by utilizing only partial knowledge of the ML model. To evaluate our attack, we consider a SHS correlating a patient's vitals collected from 8 different smart medical devices and detecting 11 benign scenarios, including 6 normal and 5 disease-affected activities using different ML models. Our evaluation shows that the proposed attack can achieve 32.27\% accuracy drop and 20.43\% success rate in untargeted and targeted attacks, respectively, leading to misclassification of the normal user activities and disease-affected scenarios in a SHS.

\vspace{6pt}
\noindent\textit{Contributions:} In summary, our contributions are three-fold: 

\begin{itemize} [nosep, wide=0pt, leftmargin=*]
    \item We present an adversarial ML-based, data-driven attack to identify flaws in the underlying ML model in a SHS. 
    We consider two different attacks: training phase attack (poisoning) and testing phase attack (evasion). 
    \item We implemented four white-box attacks and one black-box attack utilizing 5 different adversarial ML algorithms.
    \item We evaluated our proposed attack against a ML-based SHS consisting of 8 different smart medical devices. Our extensive evaluation illustrates that it can successfully downgrade the accuracy of a ML model in a SHS.
\end{itemize}

\noindent\textit{Organization:} 
We provide an overview of adversarial attacks in healthcare systems in Section~\ref{sec:related_work}. The detailed overview of SHS, as well as the adversarial ML, is provided in Section~\ref{sec:background}. In Section~\ref{sec:threat_model}, we explain our threat model considering adversary goals, capabilities, and methodology. We explore the feasibility of our attack in the case of various SHS scenarios in Section~\ref{sec:performance_evaluation}. We conclude the paper in Section~\ref{sec:conclusion}. 

\section{Related Work} \label{sec:related_work}

In this section, we discuss existing adversarial attacks to SHSs and explain how our attack model is different from the existing ones.

In recent years, several adversarial attacks on ML models have been reported by the researchers where the adversaries manipulate the input data distribution to cause incorrect classification in the output. Most of the adversarial attacks target the input data distribution to alter the training data and misclassify the output~\cite{mozaffari2014systematic}, \cite{hayes2019logan}. Again, different ML models (e.g., Neural networks, deep neural networks, etc.) are exposed to small modifications of the input data during test time which can be used by the adversaries to manipulate the data distribution by enforcing poisoned test data~\cite{biggio2013evasion}. In ML-based healthcare systems, adversaries mostly try to alter the data distribution in multi-layer ML classifier to change the predicted disease~\cite{finlayson2019adversarial}. Most of the attacks in healthcare domain target medical imaging data to alter the predicted disease. Mirsky et al. showed how an attacker could use a 3D conditional Generative Adversarial Network (GAN) to add or remove evidence of medical conditions from volumetric (3D) medical scans~\cite{mirsky2019ct}. Taghanaki et al. tested two state-of-the-art Deep Neural Network (DNN) for chest X-ray images against ten different adversarial attacks and showed that DNN in medical image classification is vulnerable against adversarial attacks~\cite{taghanaki2018vulnerability}. Kim et al. presented how a subtle universal adversarial perturbations can be added to a medical image to alter the predicted labels with high confidence~\cite{kim2019exploiting}. In a recent work, researchers proposed a novel approach for generating adversarial examples to attack Convolutional Neural Network (CNN) based segmentation models for medical images~\cite{chen2019intelligent}. Another type of adversarial attacks can be executed in Electronic Health Records (EHR) to change the future treatment plan of a patient. Choi et al. used GAN to learn the distribution of real-world multi-label discrete EHR and generate adversarial patient records~\cite{choi2017generating}. Sun et al. presented a framework that identifies the susceptible locations in a time sequence medical records utilizing adversarial attacks on deep predictive models~\cite{sun2018identify}. 

\noindent \textbf{\textit{Difference with Existing Attacks:}}
Our threat model is an entirely new approach to attack a ML-based SHS. While most of the works consider single medical device data or stored healthcare data to apply adversarial attacks, in our work, we consider a connected multi-device SHS utilizing ML models for disease and normal activity identification. The main differences between existing attacks and our approach can be noted as follows: (1) Most of the prior works focus on manipulating data distribution of a ML model in offline mode. Here, we consider a realtime ML-based SHS and manipulate the input data distribution of different medical devices to perform the adversarial attacks. (2) While the existing attacks focus on generating multi-label discrete EHR \cite{choi2017generating}, we detect the minimum number of the compromised medical devices in a SHS to initiate an attack. (3) Rather than depending on training model and full knowledge of the data distribution of ML algorithm, we consider the threshold of data injection for a successful attack using partial knowledge of the data range~\cite{hayes2019logan}. (4) Different from identifying susceptible locations in offline medical records, our attack finds the critical devices in realtime SHS to perform adversarial attacks which has higher impact on the patient's status and overall treatment~\cite{sun2018identify}. (5) Unlike adversarial attacks on medical images \cite{taghanaki2018vulnerability}, \cite{finlayson2019adversarial, kim2019exploiting, chen2019intelligent}, we consider a ML-based SHS as our attack target/platform.

\section{Background}\label{sec:background}
In this section, we discuss different components of SHS and different design assumptions that we have considered in our attack. In addition, we discuss the impact of adversarial ML in the context of a SHS.

\vspace{-0.2cm}
\subsection{Connected Smart Healthcare System Overview}

\begin{figure}[t!]
\vspace{-0.2cm}
        \centering
        \includegraphics[width=0.8\linewidth]{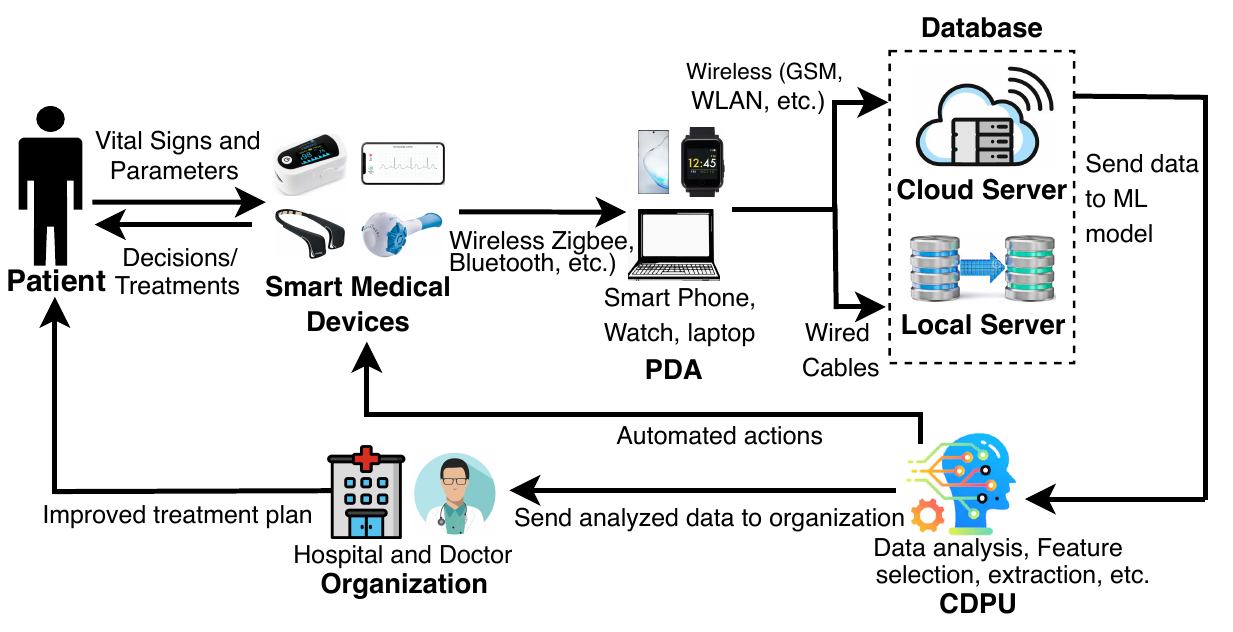}
        \vspace{-0.1cm}
        \caption{An example of a smart connected healthcare system.}
        \label{fig:background}
        \vspace{-0.6cm}
\end{figure}

A SHS can consist of single or a group of smart medical devices (e.g., wearables, wireless devices, implantable devices, etc.) to collect data from a patient's body to provide improved treatments and realtime monitoring \cite{newaz2019healthguard}. SHS considers different medical (patient's vitals) and non-medical parameters (e.g., physical status, location of the patient, etc.) to understand the overall condition of the patient and provides realtime monitoring. Smart medical devices take the vital signs as an analog signal, convert them to a digital signal and send information as a network packet to a Personal Digital Assistant (PDA) via wireless technology (Bluetooth, Zigbee, etc.), which can consist of a smartphone, a laptop, or a smartwatch. PDA works as a user interface and forwards data to a database (e.g., cloud server, local server, etc.). The database sends data to a Central Data Processing Unit (CDPU), which uses a ML model to select and extract features from the dataset. CDPU runs the ML model to detect patient disease status, normal activity of a patient, and threats in a SHS. Next, it takes automated actions (e.g., pushing a new dose of medicine, change of medication, etc.) to provide improved treatment for the patient and sends analyzed data to the authorized entity (hospital or doctor). Finally, the doctor sends a notification to the patient with updated treatment plans for his health status. Figure~\ref{fig:background} shows an example of SHS with several smart medical devices (e.g., EEG, ECG, pulse oximeter, etc.) connected to a patient's body to collect data for different vital signs (e.g., blood oxygen, neural activities, heart rate, etc.).

\vspace{-0.2cm}

\subsection{Adversarial ML in a SHS}


To understand how adversarial ML works in a SHS, we assume a ML model M is trained over input pair (X,y) from the data distribution $\mu$ with a randomized training procedure of randomness r (e.g., random weight initialization, dropout, etc.). Here, inputs (X,y) are smart medical devices data that are collected and preprocessed before running into the model. The model parameters $\theta$ is represented as follows: $\theta \leftarrow train(M, X, y,r)$. In a white-box attack on this ML model, an adversary has complete knowledge about the model (M) used for classification, and the algorithm used in training. He has access to the training data distribution ($\mu$) and is able to identify the feature space where the model is vulnerable. In a black-box attack, an adversary does not have knowledge about the model but uses information about the settings or past inputs to analyze the vulnerability of the ML model.

\section{Threat Model}\label{sec:threat_model}


In this section, we discuss the adversarial goals and attackers capabilities considered in our proposed attack.  We also describe our attack assumptions focusing on a use case scenario and explain the attack methodology in a ML-based SHS.

\vspace{-0.1cm}
\subsection{Adversarial Goals}

An adversary attempts to provide an input to a ML-based SHS to alternate the decision of the ML classifiers and manipulate the automated decisions of the SHS. We categorize the goals of an adversary in the following three categories based on the impact on the classifier output integrity:

\begin{itemize}[nosep, wide=0pt, leftmargin=*]

    \item Untargeted Attack (UA): An adversary tries to alter the output of the ML classifiers of a SHS from the actual output. For example, a high blood pressure patient will be predicted to a different disease resulting in a malicious treatment plan.
    
    \item Targeted Attack (TA): An adversary attempts to alter the input data of the ML classifier to change the output to a specific class. For instance, a high blood pressure patient's data is manipulated in such a way that it is predicted as a specific condition of abnormal oxygen levels.
    

    \item Targeted Device Attack: The adversary tries to find the minimum number of devices needed to be compromised to launch an attack in a ML-based SHS. For instance, to change the status of a stressed patient to a heart attack, it may only need to compromise the oxygen saturation device.
    
\end{itemize}
\vspace{-0.1cm}

\vspace{-0.1cm}
\subsection{Adversarial Capabilities}\label{cap}

\begin{figure}[h!]
\vspace{-0.2cm}
        \centering
        \includegraphics[width=\linewidth, height=3.5cm]{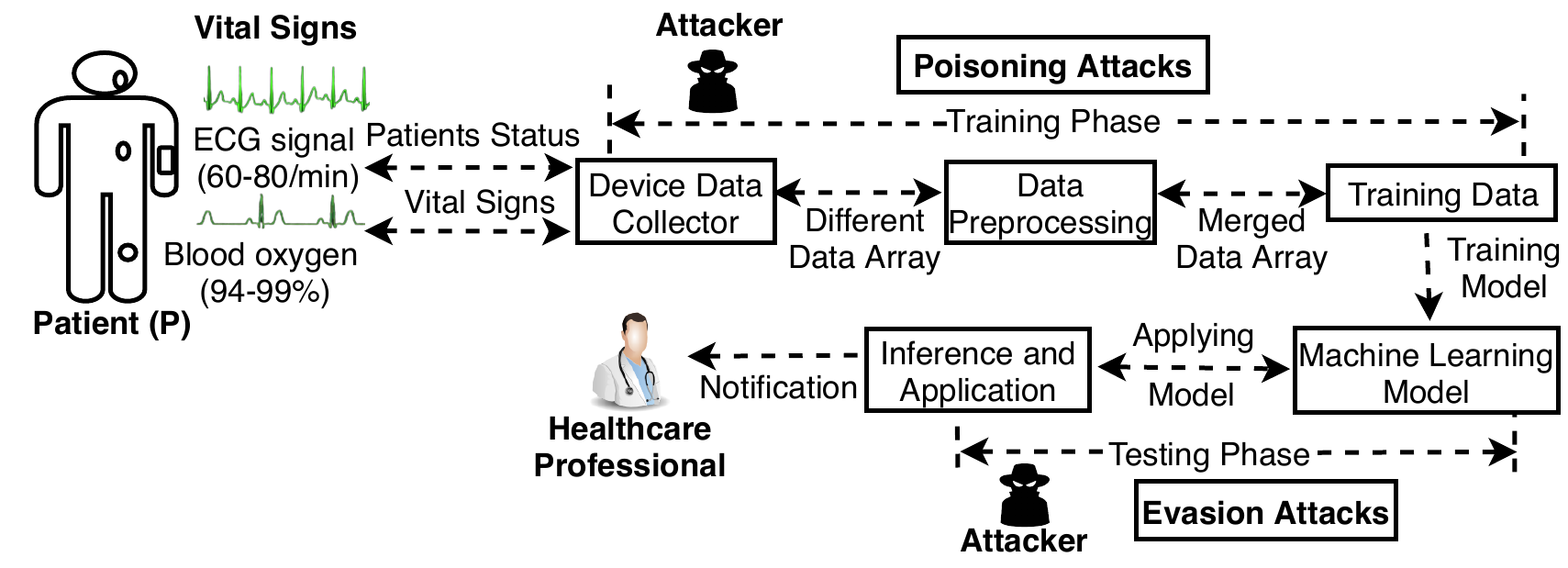}
        \caption{An example of adversarial attack in a SHS.}
        \label{fig:attack}
        \vspace{-0.5cm}
\end{figure}

To understand the attack scope of our work, we assume a patient (P) admits to a hospital having breathing problems (abnormal oxygen level) (Figure~\ref{fig:attack}). For emergency monitoring, a SHS consisting of an ECG device, a sphygmomanometer, a pulse oximeter, and an EEG is placed on P for cardiac, blood pressure, oxygen level, and neurological activity monitoring, respectively. This SHS uses a ML-based classifier to predict the disease and perform realtime monitoring of the patient. We assume that the devices are working correctly, and no compromised device is in the system. Here, we introduce a new adversarial attack that can exploit a SHS and change the patient's disease status to provide the wrong treatment.  To perform the adversarial attack on the ML model, we consider following capabilities for the adversary:

\begin{itemize}[nosep, wide=0pt, leftmargin=*]

    \item Data Distribution: As different smart medical devices generate different ranges of data, an adversary may know partial data distribution of the devices. Based on this, it may manipulate a data value within a certain threshold to change patient disease status or normal activity pattern.

    \item SHS Architecture: An adversary may know the complete or partial architecture of the SHS including number of devices, device correlation, etc. to perform an attack. 

    \item Output Label: An adversary may know the output labels (e.g., disease states and normal activity pattern) of the ML model to initiate an attack. 

    \item ML Model: Different ML models (e.g., Decision Tree, Random Forest, etc.) are used for disease and user activity detection. An adversary may have the knowledge of underlying ML model used to classify the patient status.

\end{itemize}

\subsection{Attack Methodology}

A ML-based SHS can be considered as a data processing pipeline where the vital signs collected from the patients are analyzed to diagnose the disease and provide necessary treatment. A primary sequence of operations of the SHS is illustrated in Figure~\ref{fig:attack}. A device data collector module collects data representing patient's vital signs and statuses from different smart medical devices and forwards them to the data pre-processing module. A data preprocessing module samples the data according to the corresponding sampling frequencies and saves them as an array. For example, a heart rate monitoring device monitors the heart rate of a patient in a minute, whereas the ECG device monitors the cardio-vascular state of a patient in every 10 seconds. The sampled data are used to train the ML model of the SHS for disease detection and realtime monitoring. The training data are labeled with different diseases and benign states (e.g., high blood pressure, low sugar, etc.) to understand the data pattern for different scenarios. In the testing phase, physiological data collected from the patients are analyzed based on the previously trained ML model to detect different diseases or benign states of the patient. Here, our attack methodology can be defined in the context of the data processing pipeline. An adversary attempts to manipulate either the collection or the processing of data to corrupt the ML model, hence altering the original output. We divide the attack into the following two categories in a SHS:


\textbf{Poisoning Attack:} This type of attack takes place during the training time of ML in a SHS (Figure~\ref{fig:attack}). In this attack, an adversary carefully manipulates the training data to compromise the whole learning process. As explained in Section~\ref{cap}, an adversary may know the data distribution of the ML model and can change the value of the input data to a certain threshold. Here, the adversary can use data injection, modification, and logic corruption methods to manipulate the training data. In data injection, an adversary does not have any access to the training data and algorithm but can augment a new data to the training set. In data modification, an adversary poisons the training data directly by modifying the data before it is used for training the target model. In logic corruption, an adversary has the ability to tamper the ML model. Using these capabilities, an adversary can affect the overall learning process of the ML model of the the SHS to misdiagnosed the test data which could lead to mistreatment of a patient. 
\begin{table*}[t!]
\caption{Devices and parameters considered for monitoring health condition~\cite{newaz2019healthguard, diabetes, terzano2002atlas, ashrafi2015modified}.}
\vspace{-0.1cm}
    \label{table:devices}
\centering
\fontsize{14}{16}\selectfont
\resizebox{1\textwidth}{!}{
\begin{tabular}{|l|l|p{8cm}|l|}
\hline
\multicolumn{1}{|c|}{\textbf{Device Monitoring Type}} & \multicolumn{1}{c|}{\textbf{Model}}       & \multicolumn{1}{c|}{\textbf{Feature Parameter Value}}                                & \multicolumn{1}{c|}{\textbf{Database}}            \\ \hline
Heart Rate and Blood Pressure                                            & QuadioArm                                 & 60-100 beats per minute, Systolic (120 mm Hg) and Diastolic (80 mm Hg)                                                                & Fetal ECG Synthetic Database, Data.Gov                                                      \\ \hline
Blood Glucose                                         & MiniMed™ 670G Insulin Pump System         & 70-130 mg/dl                                                                         & UCI ML Database of Diabetes                                           \\ \hline
Blood Oxygen                                          & iHealth Air Wireless Pulse Oximeter       & Oxygen Saturation Level $\ge$ 94\%                                                   & Pattern Analysis of Oxygen Saturation Variability                                 \\ \hline
Respiratory and Sweating Rate                                      & QuardioCore                               & 12-20 Breaths per minute, $0.5\mu$/min/$cm^2$                                                             & BIDMC PPG and Respiration Dataset                                               \\ \hline
Blood Alcohol                                         & Scram Continuous Alcohol Monitoring (Cam) & 0.08 g/dl                                                                            & StatCrunch Dataset                                                                 \\ \hline
Blood Hemoglobin                                      & Germaine AimStrip Hb Hemoglobin Meter     & 12.3-17.5 g/dl                                                                       & Hemoglobin Data in DHS Survey                                                 \\ \hline
Neural Activity                                       & Emotiv Insight                            & Delta (0.5-4 Hz), Theta (4-8 Hz), Alpha (8-12 Hz), Beta (16-24 Hz) \& EEG data & EEG Data                                                                  \\ \hline
Sleep and Human Motion                                                 & Fitbit Versa Smart Watch                  & REM and NREM sleep cycle                                                             & The CAP Sleep Database                                                         \\ \hline
\end{tabular}}
    \vspace{-0.5cm}
\end{table*}

\textbf{Evasion Attack:} In evasion attack, adversary tries to deceive the SHS by enforcing adversarial samples during the testing phase. An adversary does not have any influence over the training data but can access the ML model to obtain sufficient information. As a result, it attacks the ML model and manipulates the model to misclassify the patient status in a SHS. Generally, evasion attacks are classified in two categories: \textit{white-box attack} and \textit{black-box attack}. 
    
\noindent (1)~\textit{White-Box Attacks:} In a white-box attack, an adversary has complete knowledge about the ML model used in a SHS (e.g., type of neural network along with the number of layers). The adversary knows the algorithm used in the training phase and can access the training data distribution. Moreover, it knows the parameters of fully trained architecture. We perform HopSkipJump, Fast gradient method, Carlini \& Wagner, and Decision tree-based attack in the ML model of a SHS.
        
    \begin{itemize} [nosep, wide=0pt, leftmargin=*]
        
        \item HopSkipJump Attack: HopSkipJump is an efficient query algorithm and a powerful decision-based generator of adversarial examples which utilizes distance-vector as a hyper-parameter~\cite{klove2010permutation}. In our attack, we obtained optimal solution by using Chebyshev distance, which calculates the largest magnitude among each element of a vector to determine the distance between adversarial and the original samples. 


        \item Fast Gradient Method (FGM) Attack: The fast gradient method uses the gradient of the underlying model to find adversarial examples~\cite{goodfellow2014explaining}. The original input is manipulated by adding or subtracting a small error in the direction of the gradient with the intent to change the behavior of the learning model. In our model, we considered an attacker’s capability (threshold) as adding a small error in the direction of gradient to temper classification of the model. 
        
            
        \item Carlini \& Wagner (C\&W) Attack: This is the state-of-the-art white-box attack in which targeted adversarial attacks are considered as an optimization problem to take advantage of the internal configuration of ML model~\cite{croce2019scaling}. In our model, we found Euclidean distance to be optimal for computing the difference between adversarial and original examples. 
            
        \item Crafting Decision Tree Attack: Here, an adversary exploits the underlying tree structure of the decision tree classifier model~\cite{papernot2016transferability}. For a given sample and a tree, an adversary searches for leaves with different classes in the neighborhood of the leaf corresponding to the decision tree’s original prediction for the sample.We performed this attack for enumerating the minimum number of devices that need to be compromised for initiating an attack.
        
        \end{itemize}

\noindent (2)~\textit{Black-Box Attack:} In a black-box attack, an adversary does not know the ML model in the SHS and uses the information about settings or past inputs to analyze the vulnerability of the model. We perform the Zeroth Order Optimization attack in a SHS as a black-box attack.
        
    \begin{itemize}[nosep, wide=0pt, leftmargin=*]
        
        \item Zeroth Order Optimization (ZOO) Attack: ZOO attack is based on a coordinate descent method using only the zeroth-order oracle (without gradient information), which can effectively attack the black-box ML model. This is called a state-of-the-art black-box attack \cite{chen2017zoo}. 

\end{itemize}

\section{Exploring Various Attack Scenarios}\label{sec:performance_evaluation}

To demonstrate the feasibility of our proposed attack, we performed the poisoning and evasion attacks successfully against a ML-based SHS. We consider several research questions to evaluate our proposed adversarial attack.

\begin{itemize}[nosep, wide=-10pt, leftmargin=*]

   \item[] \textbf{RQ1} What is the impact of proposed attack after performing poisoning attack in a SHS? (Sec~\ref{poison})
    \item[]\textbf{RQ2} What is the impact of our attack after performing untargeted attack to find most significant devices? (Sec~\ref{evasion:decision})
     \item[]\textbf{RQ3} What is the impact of proposed white-box and black-box attacks on underlying ML model? (Sec~\ref{evasion:different_attacks})
     \item[]\textbf{RQ4} What is the performance of ML model in a SHS after performing threshold-based attacks? (Sec~\ref{evasion:thresholds})

\end{itemize}

\begin{table}[htbp]
\vspace{-0.2cm}
\caption{Device-activity correlation in SHS~\cite{newaz2019healthguard,sikder2019aegis, sikder2019context}}
\vspace{-0.1cm}
    \label{table:disease}
\centering
\fontsize{24}{28}\selectfont
\resizebox{\columnwidth}{!}{
\begin{tabular}{ll|l|l|l|l|l|l|l|l|l|l|l|}
\cline{3-13}
\multicolumn{2}{l|}{}                                                                                                               & \textbf{ECG}              & \textbf{SW}               & \textbf{BP}               & \textbf{GL}               & \textbf{BR}               & \textbf{OX}               & \textbf{SM}               & \textbf{HG}               & \textbf{AL}               & \textbf{NA}               & \textbf{HM}                                         \\ \hline
\multicolumn{1}{|l|}{\multirow{5}{*}{\textbf{\begin{tabular}[c]{@{}l@{}}Disease\\  Type\end{tabular}}}}  & \textbf{High Blood Pressure}     & -                         & \checkmark & \checkmark & \checkmark & -                         & \checkmark & \checkmark & \checkmark & \checkmark & \checkmark & -                          \\  
\multicolumn{1}{|l|}{}                                                                                   & \textbf{High Cholesterol}        & -                         & \checkmark & \checkmark & \checkmark & -                         & \checkmark & -                         & \checkmark & -                         & \checkmark & -                                        \\  
\multicolumn{1}{|l|}{}                                                                                   & \textbf{Excessive Sweating}      & \checkmark & \checkmark & \checkmark & \checkmark & -                         & \checkmark & -                         & \checkmark & -                         & \checkmark & \checkmark             \\  
\multicolumn{1}{|l|}{}                                                                                   & \textbf{Abnormal Oxygen Level}   & \checkmark & -                         & \checkmark & \checkmark & \checkmark & \checkmark & \checkmark & -                         & -                         & \checkmark & \checkmark                   \\  
\multicolumn{1}{|l|}{}                                                                                   & \textbf{High or Low Blood Sugar} & \checkmark & \checkmark & \checkmark & \checkmark & -                         & \checkmark & -                         & \checkmark & -                         & \checkmark & -                                          \\ \hline
\multicolumn{1}{|l|}{\multirow{6}{*}{\textbf{\begin{tabular}[c]{@{}l@{}}Activity \\ Type\end{tabular}}}} & \textbf{Sleeping}                & \checkmark & -                         & \checkmark & \checkmark & \checkmark & \checkmark & -                         & -                         & -                         & -                         & -                                            \\  
\multicolumn{1}{|l|}{}                                                                                   & \textbf{Walking}                 & \checkmark & \checkmark & -                         & \checkmark & \checkmark & \checkmark & -                         & \checkmark & -                         & \checkmark & \checkmark                 \\  
\multicolumn{1}{|l|}{}                                                                                   & \textbf{Stress}                  & \checkmark & \checkmark & \checkmark & -                         & \checkmark & -                         & -                         & -                         & -                         & \checkmark & -                                     \\  
\multicolumn{1}{|l|}{}                                                                                   & \textbf{Exercise}                & \checkmark & \checkmark & \checkmark & \checkmark & \checkmark & \checkmark & -                         & -                         & -                         & \checkmark & \checkmark                   \\  
\multicolumn{1}{|l|}{}                                                                                   & \textbf{Heart-Attack}            & \checkmark & \checkmark & -                         & -                         & \checkmark & -                         & -                         & -                         & -                         & \checkmark & -                                            \\ 
\multicolumn{1}{|l|}{}                                                                                   & \textbf{Stroke}                  & \checkmark & -                         & \checkmark & -                         & -                         & -                         & -                         & \checkmark & -                         & \checkmark & \checkmark                 \\ \hline
\end{tabular}}

\vspace{-0.1cm}
\end{table}


\subsection{Environment Setup and Methodology}
\begin{figure}[h!]
        \centering
        \includegraphics[scale=.15]{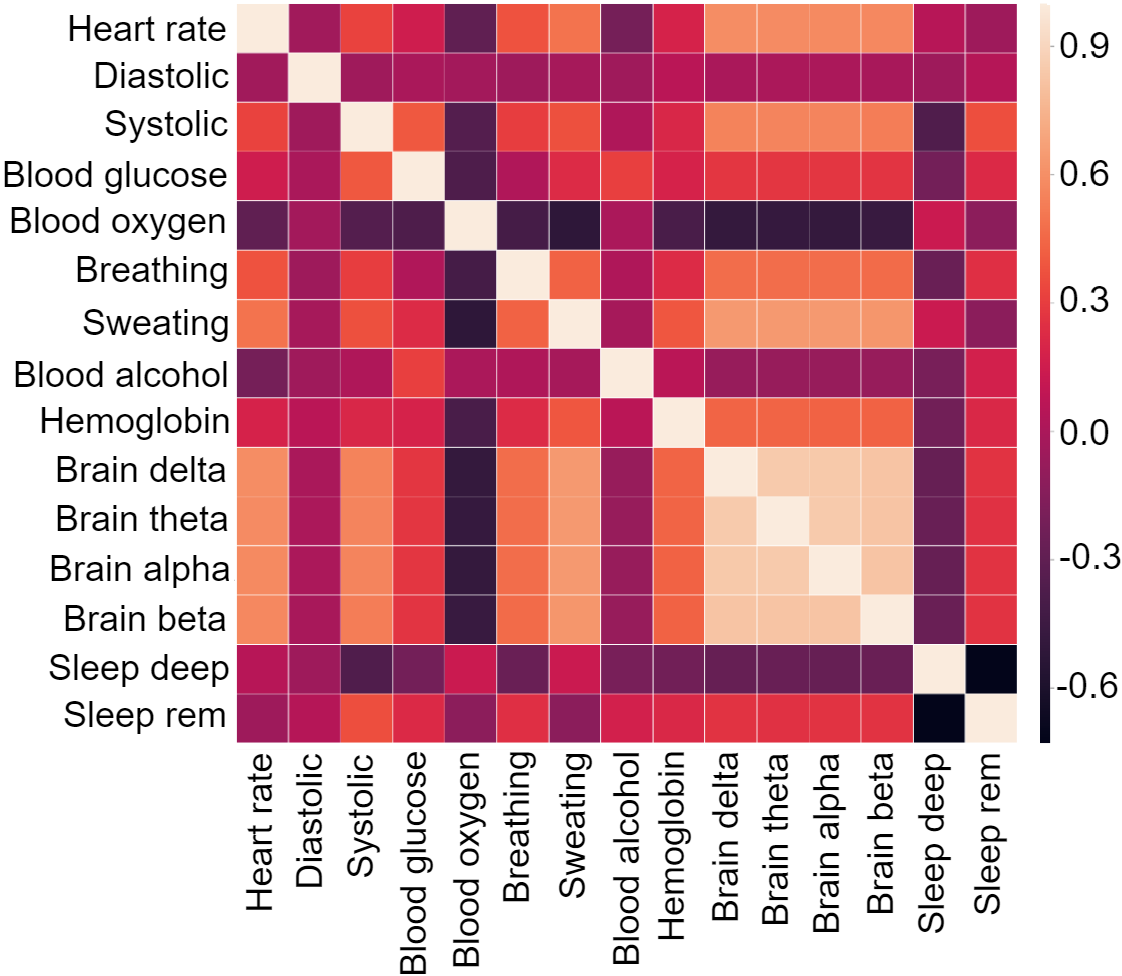}
        \vspace{-2pt}
        \caption{Correlation matrix of selected features used in SHS.} 
        \label{fig:corr}
        \vspace{-0.3in}
\end{figure}

We consider a SHS consisting of eight different devices detecting five different disease states and six different regular activities. We collected data from eight different smart medical devices available on various public healthcare databases that gave us ten vital signs: the heart rate (ECG), blood pressure (BP), blood glucose (GL), oxygen (OX) saturation, blood hemoglobin (HG), breathing rate (BR), alcohol level (AL), neural activity (NA), human motion (HM) and sleep monitoring (SM) (Table~\ref{table:devices}). These vital signs are represented by fifteen different features and the correlation among these features is presented in figure~\ref{fig:corr}. We considered the ranges 
of different vital signs of humans (e.g., heart rate, blood pressure, etc.) as the normal state. Our considered SHS uses different ML algorithms to detect five different disease scenarios (high blood pressure, high cholesterol, excessive sweating (SW), abnormal oxygen level, abnormal blood sugar) and six regular user activities (sleeping, walking, exercise, stress, heart attack, and stroke situation) based on the collected device data. The device-activity correlation is shown in Table~\ref{table:disease}. 

As the underlying ML model of the SHS, we considered four different ML algorithms (Random Forest (RF), Decision Tree (DT), Artificial Neural Network (ANN), Logistic Regression (LR)) because of multi-class classification and hierarchical patterns in our dataset. We collected 17,000 data instances for healthy and disease-infected people and divided our data into two sections, where 70\% of the collected dataset was used to train the ML model, and 30\% of the collected data was used for testing~\cite{amazon}. We executed the attack in this SHS by simulating evasion and poisoning attacks based on the attack methodology described in Section~\ref{sec:threat_model}. For a poisoning attack, we considered that an adversary poisoned the training data while he/she had the access to the ML model. Here, an adversary manipulates the output status of the patient to alter the detected disease or normal user activity could be manipulated into a disease state. To perform the evasion attack, we considered that an adversary adjusted malicious samples during the testing phase. For evasion attacks, we also conducted white-box and black-box attacks by performing gradient-based, decision boundary-based, and coordinate-based attacks to the ML algorithms of the SHS. We built a simulation environment for SHS in \textit{MATLAB} using digital signal processing toolbox and adversarial robustness toolbox \cite{ibm} (Python library) for simulating different white-box and black-box attacks. We used \textit{Keras} and \textit{Scikit-Learn} library for training and testing the dataset in Google Colab platform.

\begin{table}[h!]
\vspace{-0.1cm}
\caption{Performance of data poisoning attacks.}
    \label{table:poison}
    \vspace{-0.1cm}
\centering
\fontsize{12}{14}\selectfont
\resizebox{\columnwidth}{!}{
\begin{tabular}{l|l|l|l|l|}
\cline{2-5}
                                                     & \textbf{Before Attack} & \multicolumn{3}{c|}{\textbf{Accuracy Drop}}                                 \\ \hline
\multicolumn{1}{|l|}{\textbf{Underlying Algorithms}} & \textbf{Accuracy}      & \textbf{10\% poisoning} & \textbf{20\% poisoning} & \textbf{30\% Poisoning} \\ \hline
\multicolumn{1}{|l|}{\textbf{RF}}                    & 95.37                  & 1.65                    & 2.03                    & 2.15                    \\ \hline
\multicolumn{1}{|l|}{\textbf{DT}}                    & 90.16                  & 4.31                    & 15.88                   & 27.31                   \\ \hline
\multicolumn{1}{|l|}{\textbf{ANN}}                   & 91.42                  & 10.28                   & 13.14                   & 27.32                   \\ \hline
\multicolumn{1}{|l|}{\textbf{LR}}                    & 88.11                  & 9.28                    & 18.93                   & 28.21                   \\ \hline
\end{tabular}}
\vspace{-0.3cm}
\end{table}


In the evaluation, we used accuracy drop of the SHS's ML model from the actual accuracy as our performance metric. While accuracy refers to the effectiveness of the ML model in detecting different patient's states, adversarial samples lead to a decrease in the accuracy of the ML model. For targeted attacks, we considered attacker success rate as we performed the targeted attack in a specific portion of the total dataset.

\subsection{Evaluation with manipulated training data}
\label{poison}

In a SHS, an adversary can have access to the CDPU in the training phase and inject bad data into the underlying ML model's training pool. In that case, the adversary can learn the training model and inject bad data (poisoning attack) to manipulate the decision of the SHS. In our poisoning attack, we have altered the vital status of 10\%, 20\%, and 30\% training data for compromising SHSs' model performance. Table~\ref{table:poison} presents the performance of the poisoning attack in a SHS. We used RF, DT, ANN, LR as our underlying ML algorithms for training the SHS. The SHS achieves highest accuracy of 95.37\% in detecting different patient's states using RF. After poisoning the training dataset, the accuracy of the SHS dropped 
2.15\% for 30\% data poisoning. We can also observe that our approach achieved highest accuracy drop of 18.93\% and 28.21\% for 20\% and 30\% data poisoning, respectively, if the underlying ML model is LR. In the case of ANN with 10\% data poisoning, the highest accuracy drop is 10.28\%.



\subsection{Significant device identification in different attacks}\label{evasion:decision}

A SHS uses the correlation between different medical devices to identify the different vital status of a patient. We used this correlation to perform crafting decision tree untargeted attacks to identify the most significant devices in a SHS (Table~\ref{table:compromise_device}). We considered the untargeted attack as it uses the whole training sample to generate adversarial examples. We can observe that for an untargeted attack to change the disease state from stroke to abnormal oxygen level, and hemoglobin devices are compromised. Similarly, glucose device is affected while changing disease state from high cholesterol to stroke. In summary, to perform an untargeted attack on underlying the ML model (DT) glucose, heart rate, and oxygen saturation devices are the most affected devices.

\begin{table}[h!]
\vspace{-0.2cm}
\caption{Crafting DT for untargeted attack.}
    \label{table:compromise_device}
\vspace{-0.1cm}
\centering
\fontsize{38}{46}\selectfont
\resizebox{\columnwidth}{!}{
\begin{tabular}{|l|l|l|l|l|}
\hline
\textbf{Attack}                                            & \textbf{Current State} & \textbf{Final State}  & \textbf{Affected Device} & \textbf{Device Count} \\ \hline
\multicolumn{1}{|c|}{\multirow{6}{*}{\textbf{Untargeted}}} & High Cholesterol       & Stroke                & Glucose                  & 1                     \\ \cline{2-5} 
\multicolumn{1}{|c|}{}                                     & High Blood Pressure    & Stroke                & Glucose                  & 1                     \\ \cline{2-5} 
\multicolumn{1}{|c|}{}                                     & Abnormal Oxygen Level  & Stroke                & Glucose, Heart rate      & 2                     \\ \cline{2-5}
\multicolumn{1}{|c|}{}                                     & Stroke                 & Abnormal Oxygen Level & Glucose, Hemoglobin      & 2                     \\ \cline{2-5} 
\multicolumn{1}{|c|}{}                                     & Sleeping                  & Drunk                 & Alcohol                  & 1                     \\ \cline{2-5} 
\multicolumn{1}{|c|}{}                                     & Stress                 & Heart Attack          & Blood Oxygen        & 1                     \\ \hline
\end{tabular}}
    
    \vspace{-0.5cm}
\end{table}

\begin{figure}[!htp]
\vspace{-0.5cm}
  \centering
  \subfloat[Targeted attack]{\includegraphics[width=0.47\columnwidth, height=3.2cm]{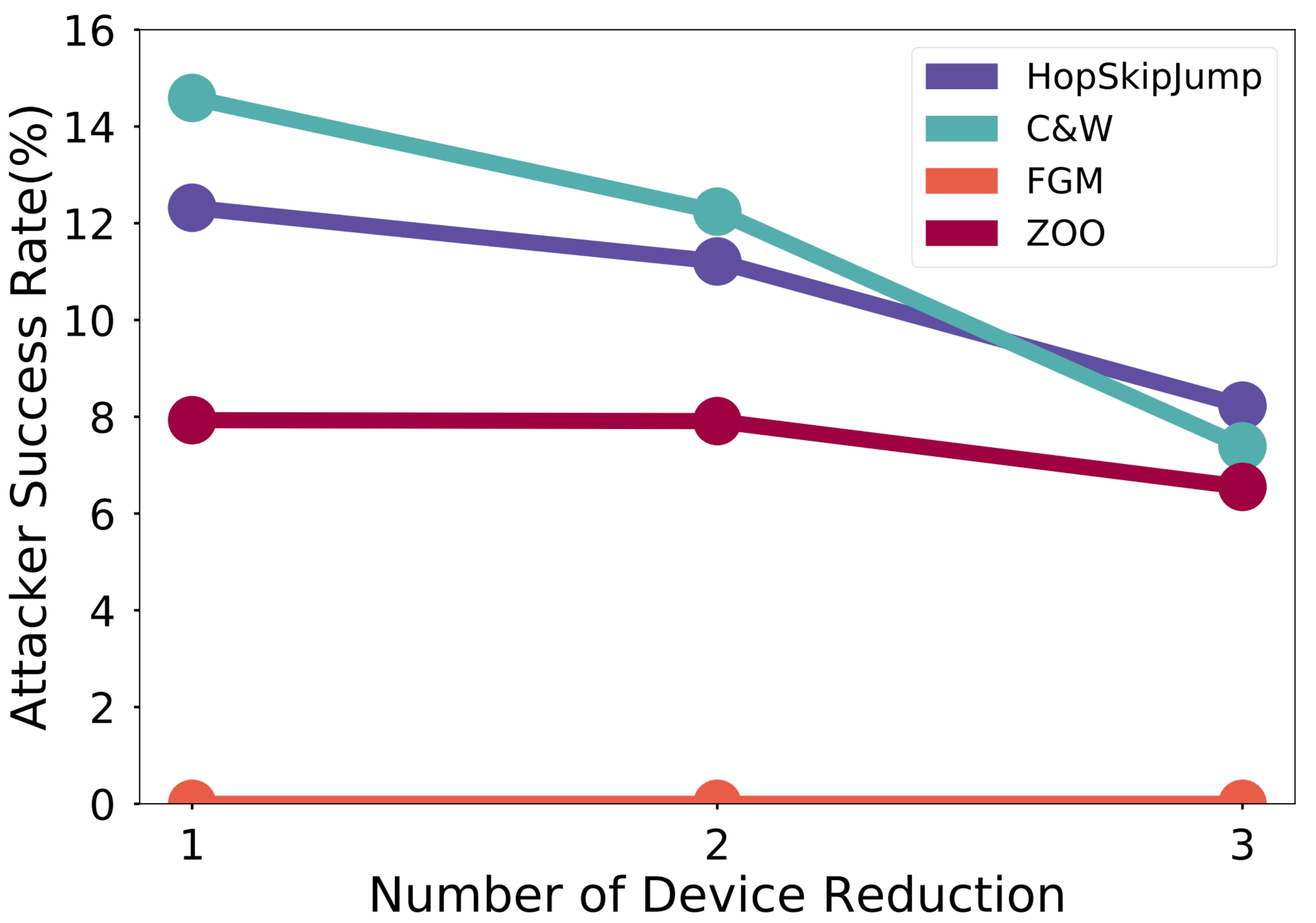}\label{fig:f1}}
    \subfloat[Untargeted attack]{\includegraphics[width=0.47\columnwidth, height=3.2cm]{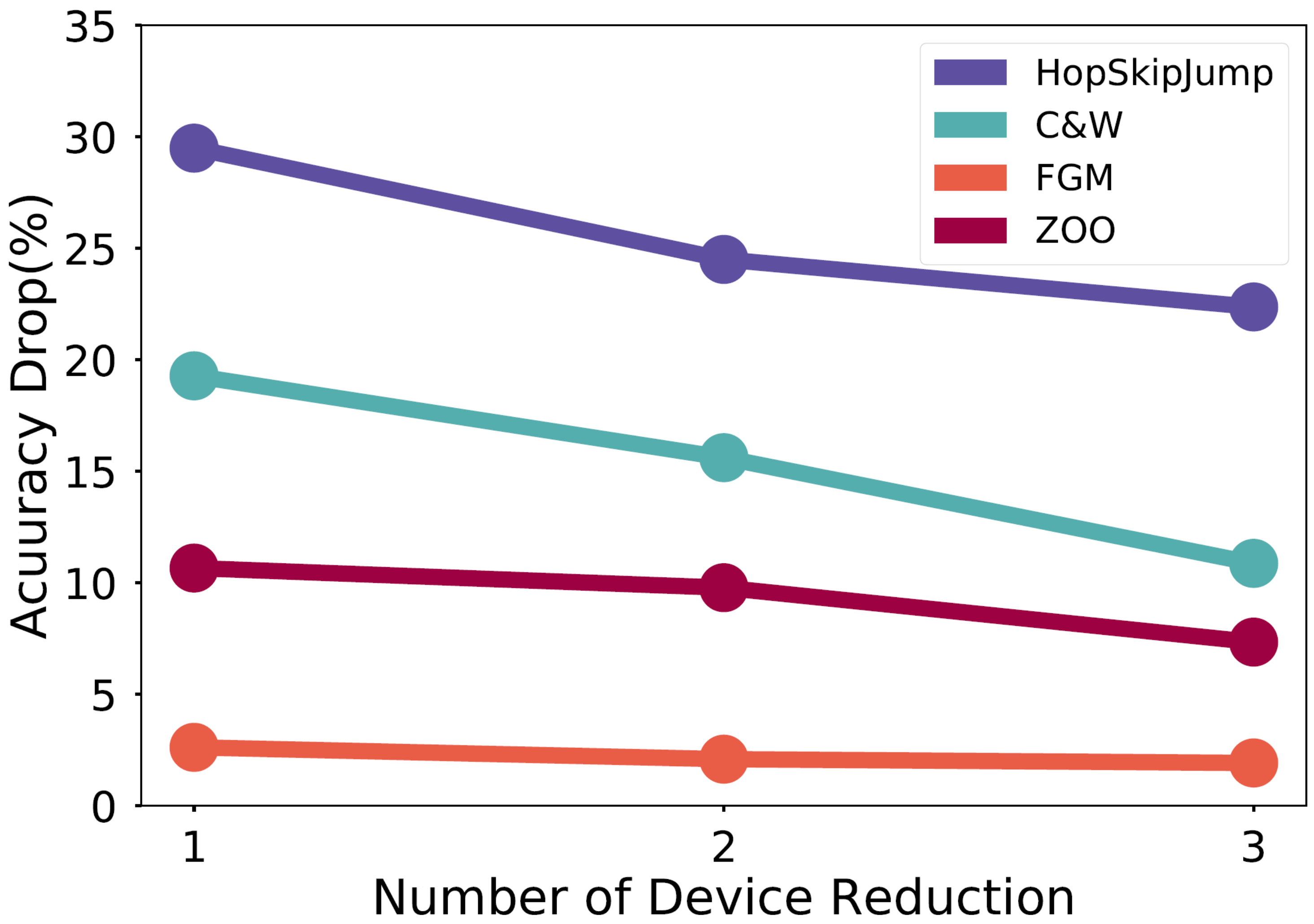}\label{fig:f2}}
\vspace{-0.1cm}
      \caption{Impact of device reduction in adversarial attacks.}
     \label{fig:accuracy}
      \vspace{-0.2cm}
\end{figure}


Figure~\ref{fig:f1} illustrates the impact of device reduction in the performance of untargeted attack. We can observe reducing number of devices drops the success rate of the proposed attack significantly. For removing one device (glucose) and two devices (glucose, blood oxygen), we achieved the highest success rate of 14.59\% and 12.24\% using C\&W attack. For removing three (glucose, oxygen, heart rate) devices, HopSkipJump can achieve the highest success rate of 8.23\%. For an untargeted attack (Figure~\ref{fig:f2}), HopSkipJump and FGM achieved the highest and lowest accuracy drop respectively for removing 1, 2, and 3 devices from the SHS. In summary, performance of adversarial attacks decrease with the reduction of devices in SHS. 

\vspace{-0.3cm}

\subsection{Evaluation with Different Attack Algorithms}
\label{evasion:different_attacks}

To evaluate the performance of the proposed adversarial attack, we performed white box and black box attacks in a SHS using different adversarial algorithms. From Table~\ref{table:different_attacks}, we can observe the highest accuracy drop (32.27\%)  and success rate (15.68\%) can be achieved using HopSkipJump for untargeted attack and targeted attack respectively. On the contrary, FGM achieved the lowest accuracy drop and success rate while the underlying model was LR. For the black-box, we performed both untargeted and targeted ZOO attacks and obtained 12.29\% accuracy drop and 8.22\% success rate respectively, where the underlying model was RF.

\begin{table}[htbp]
    \caption{Performance of white-box and black-box attacks.}
    \label{table:different_attacks}
\centering
\fontsize{12}{14}\selectfont
\resizebox{\columnwidth}{!}{
\begin{tabular}{|l|l|l|l|l|l|}
\hline
\textbf{\begin{tabular}[c]{@{}l@{}}Adversary \\ Capability\end{tabular}} & \textbf{\begin{tabular}[c]{@{}l@{}}Attack \\ Algorithms\end{tabular}} & \textbf{\begin{tabular}[c]{@{}l@{}}Underlying \\ ML model\end{tabular}} & \textbf{\begin{tabular}[c]{@{}l@{}}Actual \\ Accuracy\end{tabular}} & \textbf{\begin{tabular}[c]{@{}l@{}}Accuracy \\ Drop (UA)\end{tabular}} & \textbf{\begin{tabular}[c]{@{}l@{}}Success \\ Rate (TA)\end{tabular}} \\ \hline
\multirow{3}{*}{\textbf{White-box}}                                      & \textbf{HopSkipJump}                                                  & \textbf{DT}                                                             & 90.16                                                               & 32.27                                                                  & 15.68                                                                 \\ \cline{2-6} 
                                                                         & \textbf{C\&W}                                                         & \textbf{ANN}                                                            & 85.14                                                               & 24.14                                                                  & 20.43                                                                 \\ \cline{2-6} 
                                                                         & \textbf{FGM}                                                          & \textbf{LR}                                                             & 88.11                                                               & 3.02                                                                   & 0.26                                                                  \\ \hline
\textbf{Black-box}                                                       & \textbf{ZOO}                                                          & \textbf{RF}                                                             & 90.21                                                               & 12.29                                                                  & 8.22                                                                  \\ \hline
\end{tabular}}

    \vspace{-0.5cm}

\end{table}

\begin{figure}[h!]
\vspace{-0.5cm}
  \centering
  \subfloat[Targeted attack]{\includegraphics[width=0.47\columnwidth, height=3.1cm]{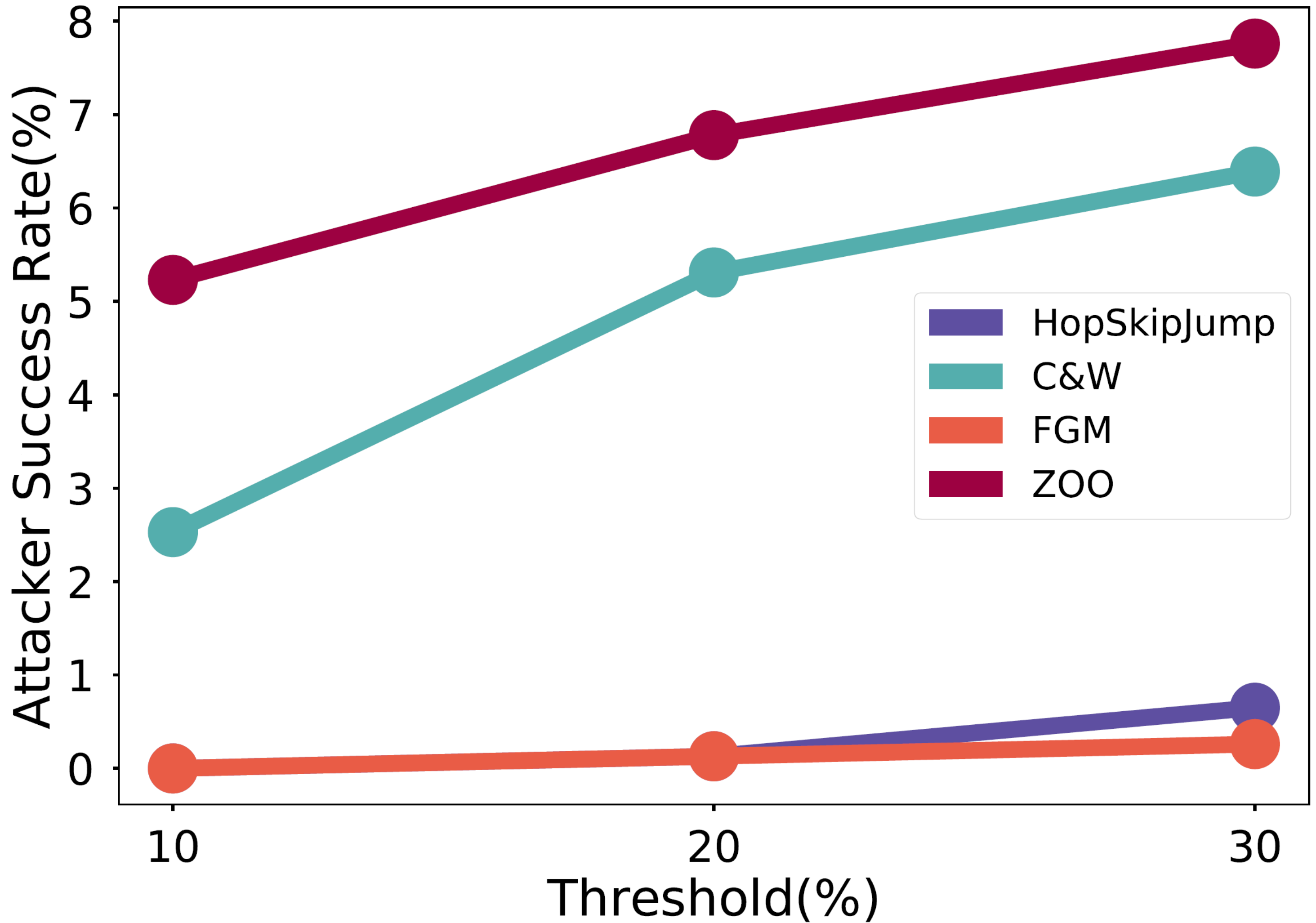}\label{fig:f3}}
\subfloat[Untargeted attack]{\includegraphics[width=0.47\columnwidth, height=3.1cm]{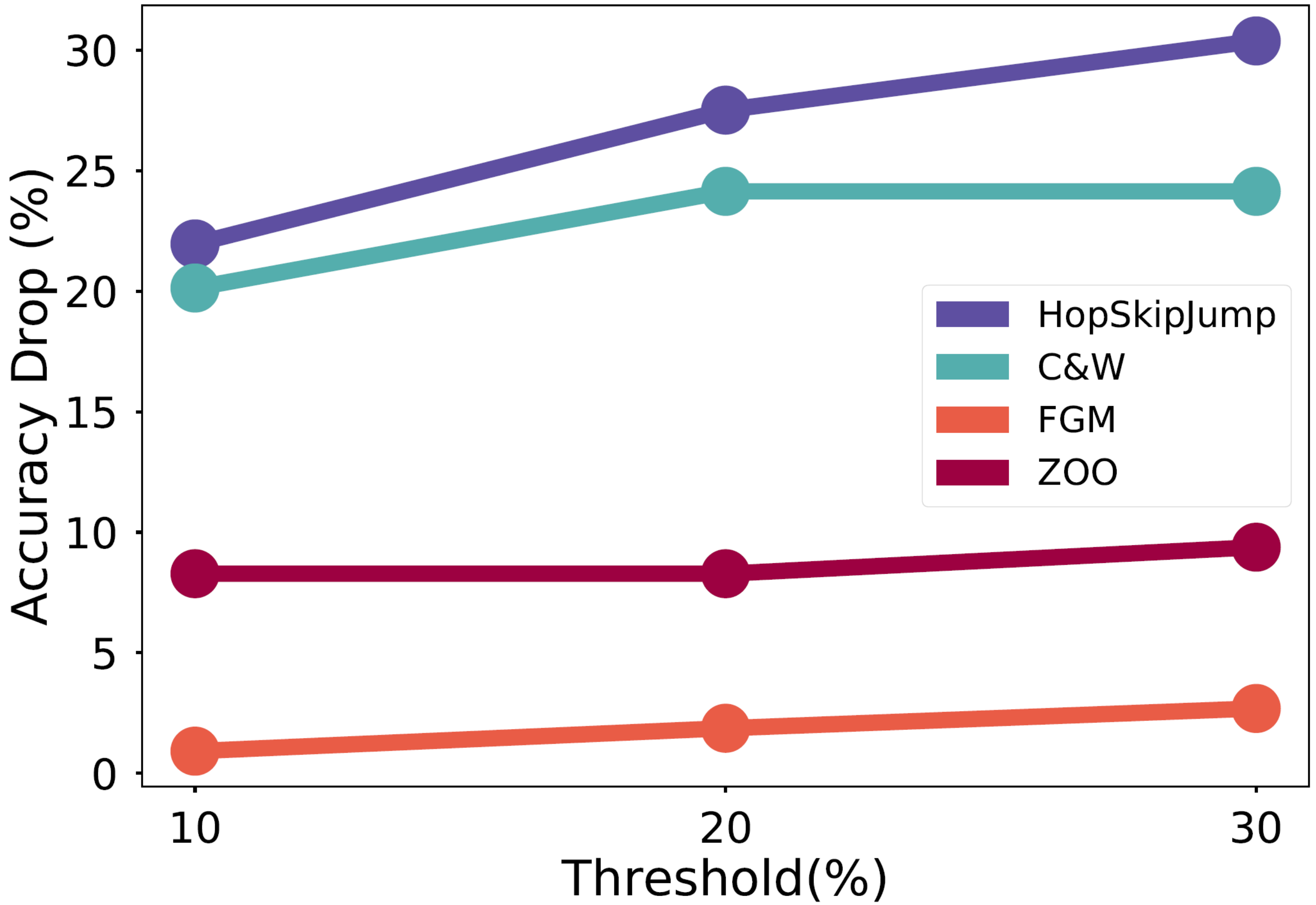}\label{fig:f4}}
\vspace{-0.1cm}
      \caption{Threshold based attacks on a SHS.}
     \label{fig:accuracy}
      \vspace{-0.5cm}
\end{figure}


\subsection{Evaluation with Considering Different Thresholds}
\label{evasion:thresholds}

We performed threshold-based attack as described in Section~\ref{sec:threat_model}. From Figure~\ref{fig:f3}, we can observe that for an targeted attack HopSkipJump and FGM failed to achieve any success rate for 10\% threshold-based attack. 
ZOO attack achieved highest success rate for all three threshold-based attacks in a SHS. During an untargeted attack (Figure~\ref{fig:f4}), ZOO attack algorithms had same accuracy drop of 8.28\% for both 10\% and 20\% thresholds. FGM had the lowest accuracy drop for all three attack. In summary, HopSkipJump achieved highest accuracy drop for 
threshold-based attacks.

\section{Conclusion}\label{sec:conclusion}

The healthcare industry is generating a vast amount of data for patients and ML is being used to analyze these data for different applications such as early disease diagnose, improved treatment plan, realtime patient monitoring, etc. However, the ML models used in different medical applications have weaknesses in terms of unpredictability and outside of the training distribution data. To address these shortcomings, in this paper, we presented 
a new adversarial ML-based attack to identify the pitfalls in the underlying ML model of a SHS. We evaluated our attack considering different adversarial settings and medical setup in a SHS. Our extensive evaluation shows that it can achieve very promising results 
for untargeted and targeted attacks 
to misclassify the normal user activities and disease affected scenarios in SHSs.


\section{Acknowledgment}\label{sec:acknowledgment}

This work was partially supported by US NSF-CAREER-CNS-1453647. 
The  views  here 
belong to 
the authors only.


\bibliographystyle{IEEEtran}
\bibliography{reference.bib}

\end{document}